# Benchmarking tree species classification from proximally-sensed laser scanning data: introducing the FOR-species20K dataset

Stefano Puliti[a †], Emily R. Lines[b], Jana Müllerová[c], Julian Frey[d], Zoe Schindler[d], Adrian Straker[e], Matthew J. Allen[b], Lukas Winiwarter[f], Nataliia Rehush[g], Hristina Hristova[g], Brent Murray[h], Kim Calders[i], Louise Terryn[i], Nicholas Coops[h], Bernhard Höfle[j], Samuli Junttila[k], Martin Krůček[l], Grzegorz Krok[m], Kamil Král[l], Shaun R. Levick[n], Linda Luck[o], Azim Missarov[l], Martin Mokroš[p], Harry J. F. Owen[b], Krzysztof Stereńczak[m], Timo P. Pitkänen[q], Nicola Puletti[r], Ninni Saarinen[k], Chris Hopkinson[s], Chiara Torresan[t], Enrico Tomelleri[u], Hannah Weiser[i], and Rasmus Astrup[a]

[a] Norwegian Institute for Bioeconomy Research (NIBIO), Division of Forest and Forest Resources, National Forest Inventory. Høgskoleveien 8, 1433 Ås, Norway
[b] Department of Geography, University of Cambridge, Cambridge, UK
[c] Department of Geoinformatics, Jan Evangelista Purkyně University in Ústí n. L., Ústí n. L., Czech Republic
[d] Chair of Forest Growth and Dendroecology, University of Freiburg, Tennenbacher Str. 4, 79106 Freiburg, Germany
[e] Georg-August-Universität Göttingen, Faculty of Forest Sciences and Forest Ecology, Burckhardt-Institute, Forest Inventory and Remote Sensing, Büsgenweg 5, 37077 Göttingen, Germany
[f] Research Unit Photogrammetry, Department of Geodesy and Geoinformation, TU Wien, Vienna, Austria
[g] Swiss Federal Institute for Forest, Snow and Landscape Research WSL, Zürcherstrasse 111, 8903 Birmensdorf, Switzerland
[h] Department of Forest Resources Management, Faculty of Forestry, University of British Columbia, 2424 Main Mall, Vancouver, British Columbia V6T 1Z4, Canada
[i] Q-ForestLab, Department of Environment, Faculty of Bioscience Engineering, Ghent University, Belgium
[j] 3DGeo Research Group, Institute of Geography, Heidelberg University, Heidelberg, Germany
[k] School of Forest Sciences, University of Eastern Finland, Yliopistokatu 7, 80101, Joensuu, Finland
[l] Department of Forest Ecology, Silva Tarouca Research Institute, Průhonice, Czech Republic
[m] Department of Geomatics, Forest Research Institute, Sękocin Stary, 3 Braci Leśnej Street, 05-090 Raszyn, Poland
[n] Land and Water, Commonwealth Scientific and Industrial Research Organization (CSIRO), Winnellie, Australia
[o] Research Institute for the Environment and Livelihoods, Charles Darwin University, Northern Territory, 0909, Australia
[p] Department of Geography , University College of London, Gower Street, London, WC1E 6BT
[q] Natural Resources Institute Finland (Luke), Latokartanonkaari 9, 00790, Helsinki, Finland
[r] CREA-FL, Council for Agricultural Research and Economics, Research Centre for Forestry and Wood, Arezzo, Italy
[s] Department of Geography, University of Lethbridge, Canada
[t] National Research Council – Institute of BioEconomy, via Biasi, 75 (38098) San Michele all'Adige (TN), Italy
[u] Faculty of Agricultural, Environmental and Food Sciences, Free University of Bolzano

## Abstract

1. Proximally-sensed laser scanning presents new opportunities for automated forest data capture and offers profound insights into forest ecosystems. However, a gap remains in automatically deriving ecologically pertinent forest information, such as tree species, without relying on additional ground data. Artificial intelligence approaches, particularly deep learning (DL), have shown promise toward automation. However, progress has been limited by the lack of large, diverse, and, most importantly, openly available labelled single tree point cloud datasets. This has hindered both 1) the robustness of the DL models across varying data types (platforms and sensors), and 2) the ability to effectively track progress in DL model development, thereby slowing the convergence towards best practice for species classification.

2. To address the above limitations, we compiled the FOR-species20K benchmark dataset, consisting of individual tree point clouds captured using proximally sensed laser scanning data from terrestrial laser scanning (TLS), mobile laser scanning (MLS) and drone laser scanning (ULS). Compiled collaboratively, the dataset includes data collected in forests mainly across Europe, covering Mediterranean, temperate and boreal biogeographic regions, and includes scattered tree data from other continents, amounting to a total of over 20,000 trees of 33 species and covering a wide range of tree size and form. Accompanying the open release of FOR-species20K, we benchmarked seven leading DL models for individual tree species classification. The benchmarked models included both point cloud (PointNet++, MinkNet, MLP-Mixer, DGCNNs) and multi-view image-based methods (SimpleView, DetailView, YOLOv5).

3. 2D Image-based models had, on average, higher overall accuracy (average OA = 0.77) than 3D point cloud-based models (average OA = 0.72). Notably, the performance was consistent (OA > 0.8) across scanning platforms and sensors, so offering versatility in deployment. Examining the results of the top-performing model (DetailView), we found that it demonstrated robustness to training data imbalances and effectively generalized across tree size.

4. The FOR-species20K dataset represents an important asset for development and benchmarking of DL models for individual tree species classification using proximally sensed laser scanning data. As such, it serves as a crucial foundation for future efforts to accurately classify and map tree species at various scales using laser scanning technology, as it provides the complete code base, dataset, and an initial baseline representative of the current state-of-the-art of point cloud tree species classification methods.

# 1- Introduction

In recent years there has been a significant push towards automating the retrieval of key forest variables from various remotely sensed data, with laser scanning technology providing the most detailed and accurate 3D information (Calders et al. 2020). Laser scanning and other 3D technologies have demonstrated exceptional capabilities in capturing forest structure, and a variety of scanning platforms exists, suited to a range of target applications. These applications range from large-scale airborne forest management inventories employing airborne laser scanning (ALS) to small-scale surveys aimed at capturing in-situ data and relying on proximally-sensed laser scanning data from various platforms, which we here define as uncrewed aerial vehicle (UAV) laser scanning (ULS), terrestrial laser scanning (TLS), and mobile laser scanning data (MLS).

While the value of laser scanning for increased understanding of the structure and function of forests has been widely acknowledged (Calders et al. 2020; Disney 2019; Krůček et al. 2019; Lines et al. 2022b; Malhi et al. 2018), its ability to routinely replace traditional ground surveys is limited by a number of factors. These of course include the cost and accessibility of sensors, but even when these are not a barrier, post-processing to retrieve interpretable information still requires significant manual work. Crucially, as identified during the recent Europa Biodiversity Observation Network (EuropaBON) online workshop on EBV workflows (Lumbierres et al. 2024), standard approaches for large-scale, reliable retrieval of ecologically important forest properties (Remotely Sensed Essential Biodiversity Variables, RS-EBV, O'Connor et al. 2015) have not yet been agreed by the community, (Lumbierres et al. 2024). The classification of tree species from laser scanning data is a crucial task for effective forest monitoring, management, and assessment of forest functions and ecosystem services Species information is key for estimating carbon storage and sequestration, growing stock volume, wood quality and properties, demography and dynamics, biodiversity, and habitat quality, properties which form the basis for successful forest management and conservation (Lines et al. 2022a). Whilst deep learning methods have shown promise for the problem of individual tree species classification (see table 1), development has largely relied on single datasets from on a small number of species in a single ecosystem type, and, to date, reliable and robust universal models have not emerged. Without reliable, robust and well-tested automatic species classification, the use of laser scanning sensors to monitor forests and their ecosystem services will always be limited by the need for such auxiliary ground data.

A review of a substantial body of literature (19 studies assessed in July 2024; see table 1) aiming to highlight the significance, feasibility, and current state-of-the-art in tree species classification using individual tree point clouds has identified the following key trends and gaps:

- **Platform variation**: The variation in platforms used to capture laser scanning data illustrates the broad spectrum of downstream applications and underlines the importance of point cloud-based tree species classification. Terrestrial datasets, particularly those using TLS data, have been among the most studied, accounting for about 45% of research. However, there is growing interest in employing high-density laser scanning data from drones or helicopters for individual tree species classification.
- **Advancements in state-of-the-art**: Over the past decade, there has been a general improvement in performance, particularly due to the advent of DL methods. These have proven superior compared to shallower methods such as Random Forests (Marinelli et al. 2022; Xi et al. 2020). Within the DL realm, PointNet++ dominates the literature in terms of the number of studies where this architecture was used. However, when comparing PointNet or PointNet++ against multi-view 2D-CNN approaches on the same dataset (Lin et al. 2023; Marinelli et al. 2022; Seidel et al. 2021), the latter generally exhibits higher accuracy (Lin et al. 2023; Marinelli et al. 2022; Seidel et al. 2021).

- **Geographic, species, and ecosystem diversity in studies**: Most studies have been conducted in mature and relatively simple forest types, predominantly in China and Finland. These typically involve managed plantation forests in temperate and boreal climates with a limited number of species. In contrast, only the studies by Liu et al. (2022a) and Allen et al. (2023) looked at more complex forest structures characterized by multiple canopy layers composed of many species with highly plastic and intermingled crowns.
- **Dataset size and diversity**: Most studies, except for the early work by Guan et al. (2015) and Hovi et al. (2016), used small datasets, averaging around 1,500 trees. These studies have also been relatively homogeneous, typically involving the classification of up to six tree species, and often relying on data from a single sensor. As a result, the models trained on these datasets are likely to have limited transferability to new data. Specifically, factors related to the scanning platform, sensor, and protocol—such as occlusion, resolution, and noise—significantly impact the data structure (Lines et al. 2022b). These dataset-specific properties may heavily influence the trained models' weights, restricting their applicability to specific forest types and data captured under specific scanning parameters and protocols.
- **Platform and sensor agnostic models**: Previous research has been constrained to single laser scanning data modalities, such as ALS, ULS, MLS, or TLS, likely due to limitations imposed by the small datasets. This has led to a significant gap in our understanding of the cross-platform performance of the different methods. Recent studies by Krisanski et al. (2021) and Wielgosz et al. (2024) have demonstrated the potential of using sensor- and platform agnostic models for forest point cloud segmentation tasks. These findings suggest that similar approaches could also be beneficial for tree species classification. The development of models that effectively transfer across different laser scanning sensors and platforms could streamline the automation of 3D forest data capture and facilitate the integration of multi-scale laser scanning datasets seamlessly. In addition, sensor- and platform- agnostic models should work on emerging sensor/platform combinations without the need for new training datasets and model training, making such models more sustainable.
- **Accessibility of data and code**: Strikingly, out of the nineteen studies reviewed, only Seidel et al. (2021) and Allen et al. (2023) made their datasets publicly available. Open DL-ready datasets expand the pool of developers and researchers who can innovate without requiring data collection, while at the same time providing transparent progress tracking. To this end, it is imperative to establish more inclusive and efficient methods for development and progress monitoring. Similarly to the data, with only few examples (Allen et al. 2023; Lin et al. 2023), code is also shared seldom.

**Table 1**. Summary of the reviewed literature in the domain of individual tree species classification using laser scanning data.

| Reference | Dataset | | | | | Methods and results | |
|---|---|---|---|---|---|---|---|
| | Platform | Country | Forest type | n trees | n species | Specific classifier | Overall accuracy |
| **Puttonen et al. (2011)** | MLS | Finland | Urban forest | 133 | 10 | SVM | 0.65 |
| **Guan et al. (2015)** | MLS | China | Urban forest | 52,013 | 10 | SVM | 0.85 |
| **Hovi et al. (2016)** | ALS | Finland | Boreal forests | 13,560 | 3 | QDA | 0.84 – 0.91 |
| **Zou et al. (2017)** | TLS | China | Plantation | NA | NA | Voxel CNN | 0.93 – 0.95 |
| **Mizoguchi et al. (2017)** | TLS | Japan | Temperate | NA | 2 | 2D CNN (bark depth images) | 0.85 – 0.91 |
| **Åkerblom et al. (2017)** | TLS | Finland | Boreal | 1,010 | 3 | KNN<br>MLR | 0.97<br>0.95 |
| | | | | | | SVM | 0.97 |
| **Terryn et al. (2020)** | TLS | UK | Temperate broadleaved | 788 | 5 | RF, KNN, MLR, SVM | 0.82 |
| **Xi et al. (2020)** | TLS | Canada and Finland | Boreal and temperate | 771 | 9 | RF,<br>Voxel CNN (ResNet-50,<br>Inception-ResNet-v2)<br>3D CNN (PointNet++ with leaf-wood information) | 0.91<br>0.86<br>0.93<br>0.96 |
| **Liu et al. (2021)** | TLS<br>ULS | China (Mongolia) | NA | 1,200 | 2 | 3D CNN (PointNet) | 0.92<br>0.89 |
| **Seidel et al. (2021) **** | TLS | Germany and US | Temperate | 690 | 7 | Multi-view 2D CNN (LeNet5; 10 images per tree)<br>3D CNN (PointNet) | 0.86<br>- |
| **Lv et al. (2021)** | ULS | China | Temperate broadleaved | NA | 4 | 3D CNN (PointNet++ with hand-crafted features) | 0.87 |
| **Chen et al. (2021)** | TLS/ULS | China | Plantation, temperate broadleaved | 1,000 | 2 | PCTSCN<br>3D CNN (PointNet)<br>3D CNN (PointNet++)<br>Voxel CNN (VoxNet)<br>2D CNN (ResNet101) | 0.89 – 0.94<br>0.83 – 0.89<br>0.89 –0.92<br>0.81 – 0.85<br>0.88 – 0.92 |
| **Liu et al. (2022a)** | MLS | China | Boreal, temperate, subtropical | 1,312 | 8 | 3D CNN (PointNet++) | >0.95 |
| **Liu et al. (2022b)** | TLS | China | Temperate | 526 | 7 | 3D CNN (PointNet)<br>3D CNN (PointNet++)<br>3D CNN (PointMLP) | 0.28<br>0.88<br>0.84 |
| **Marinelli et al. (2022)** | ALS | Italy | Mountainous temperate | 1,216 | 7 | SVM<br>3D CNN (PointNet++)<br>Multi-view 2D CNN (8 images per tree) | 0.64<br>0.67<br>0.83 |
| **Allen et al. (2023) ***** | TLS | Spain | Mediterranean | 2,478 | 5 | Multi-view 2D CNN (SimpleView; 6 images per tree) | 0.81 |
| **Fan et al. (2023)** | ALS | China | | 548 | 11 | 3D CNN (PointNet++) | 0.92 |
| **Hakula et al. (2023)** | ALS-HD | Finland | Boreal | 5,500 | 4 | RF | 86.6 |
| **Lin et al. (2023) *** | ALS | Kenya | Tropical savanna | 4,000 | 6 | Multi-view 2D CNN (SimpleView; 6 images per tree)<br>PCT | 0.7<br>0.72 |

The acronyms for the species classifiers are as follows: support vector machines (SVM), quadratic discriminant analysis (QDA), convolutional neural network (CNN), k nearest neighbour (KNN), random forest (RF), multinomial linear regression (MLR), point cloud transformer (PCT);

NA= not available;

* open code; ** open data; *** open data and code

While DL techniques show highest accuracy for automatic tree species classification at the individual tree level (Table 1), navigating the landscape of available models remains challenging, and the deployment of such models in real-world applications remains limited. One major issue is the difficulty in objectively comparing model performances, as many models are validated using internal test data, often resulting in inflated accuracy metrics that do not accurately reflect real-world conditions. In this context, the availability of public benchmark datasets becomes invaluable for the research community (Lines et al. 2022a). Open model weights and DL-ready datasets enable objective model development, advance the state-of-the-art in tree species classification, and help users understand the strengths and weaknesses of different approaches. Public leaderboards that compare model performance on benchmark datasets are essential for users looking to select the most suitable models for their needs.

This paper addresses the need for improved tree species classification by collating, standardising, and publishing a large, diverse, individual tree point cloud DL-ready benchmark dataset for platform- (airborne and terrestrial) and sensor- (consumer and survey grade) agnostic tree species classification tasks. We also establish a baseline by benchmarking current state-of-the-art DL models, underlining the importance of moving towards platform- and sensor-agnostic models. We analyse the outputs of the benchmarking exercise not only using standard ML approaches, but also by considering the context. For example, we explore the variation in performance by sensor type and species, showing the need for methods to effectively cope with data imbalances, a common challenge in nature where tree species distributions are often uneven. Finally, we publish all our data, including the test-train-validation split, and our model weights, openly with the aim of encouraging future development.

## 2 - Materials

The FOR-species20K dataset was compiled by combining 25 different datasets, composed of either open datasets or in-kind contributions from researchers, primarily in Europe but also in North America and Australia. Data were collated based on directly approaching relevant researchers, the authors' networks, and an open call disseminated via multiple channels, including academic newsletters, conference presentations and social media channels. The basic data unit is individual tree point clouds, so to ensure the quality of the dataset, we selected only submissions where the individual tree segmentation was of high quality (i.e. manual segmentation). The final number of trees included in the FOR-species20K dataset was 20,158.

### 2.1 - Data origin

The large majority (90% of the trees) of data in the FOR-species20K were collected within Europe, with additional scattered data collections from Canada, Australia, and New Zealand (see Figure 1).

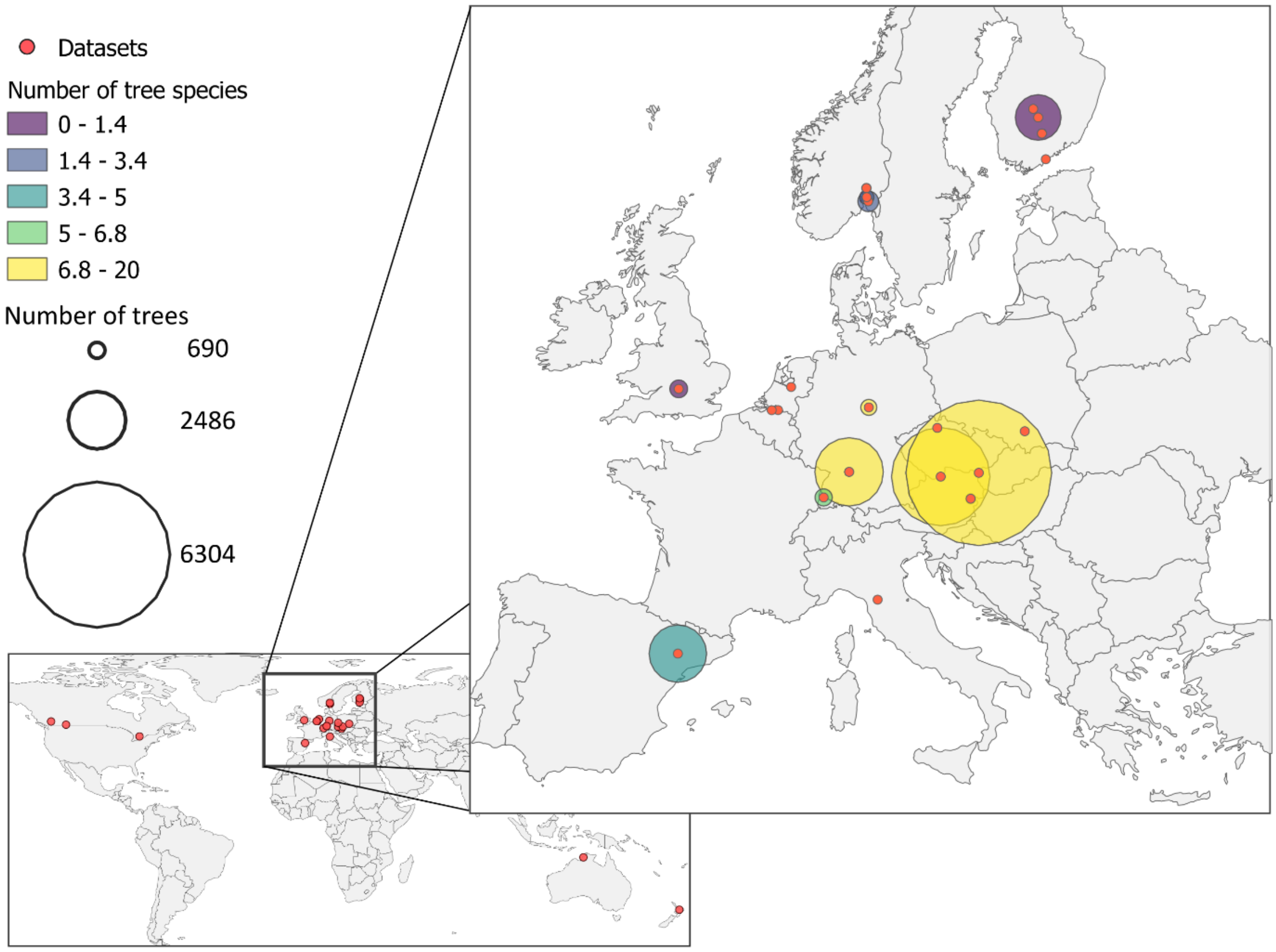

**Figure 1**. Geographic overview of the FOR-species20K dataset.

## 2.2 - Platforms and sensors

The FOR-species20K dataset is composed of mostly TLS data (70% of the trees), followed by ULS (22%), and lastly, MLS data (8%). Likely due to the longer history of forest use of TLS compared to MLS and ULS, the TLS data herein is characterized by a broader variability in terms of sensors (12 different sensors) and scanning protocols, spanning from fixed position single scan to equally spaced scans and including varying designs in between. MLS and ULS data came from only one type of sensor each. Further details on the different datasets can be found in Table 2.

**Table 2**. Summary of the individual data collections within the FOR-species20K dataset.

| Dataset name* | Reference | n trees | n species | Data type | Sensor | Biogeographic region |
|---|---|---|---|---|---|---|
| BlueCat_a | - | 3,926 | 9 | TLS | Leica P20 | temperate |
| Owen_2021 | Owen et al. (2021) | 2,478 | 5 | TLS | Leica HDS6200 | Mediterranean |
| Saarinen_2021 | Saarinen et al. (2020, 2021a, 2021b) | 1,976 | 1 | TLS | Trimble TX5 3D | boreal |
| BlueCat_b | - | 1,409 | 3 | TLS | Leica P20 | temperate |
| Calders_2022 | Calders et al. (2022) | 769 | 6 | TLS | RIEGL VZ-400 | temperate |
| Xi_2020a | Xi et al. (2020) | 661 | 3 | TLS | Optech Ilris HD | boreal |
| Seidl_2021 | Seidel et al. (2021) | 577 | 7 | TLS | Faro Focus 3D 120, Zoller and Fröhlich Imager 5006 | temperate |
| Frey_2022 | - | 472 | 6 | TLS | RIEGL VZ-400i | temperate |
| Xi_2020_2 | Xi et al. (2020) | 398 | 2 | TLS | Optech Ilris HD | temperate |
| Liang_2018 | Liang et al. (2018) | 385 | 3 | TLS | Leica HDS6100 | boreal |
| Luck | - | 352 | 1 | TLS | Leica BLK360 | tropical savana |
| Weiser_2022a | Weiser et al. (2022) | 263 | 11 | TLS | RIEGL VZ-400 | temperate |
| Luke | - | 225 | 3 | TLS | Leica P40 | boreal |
| REMBIOFOR | - | 57 | 3 | TLS | FARO Focus 3D X130 | temperate |
| Junttila | - | 51 | 1 | TLS | Leica RTC360 | boreal |
| VanDeBerge_2021 | Van Den Berge et al. (2021) | 50 | 2 | TLS | RIEGL VZ-1000 | temperate |
| Puliti_a | - | 895 | 4 | MLS | Geoslam Horizon | boreal |
| LAUTx | Tockner et al. (2022) | 434 | 6 | MLS | Geoslam Horizon | temperate |
| UBC_2022 | - | 279 | 2 | MLS | Geoslam Horizon | temperate |
| Mokros_2022 | - | 114 | 1 | MLS | Geoslam Horizon | temperate |
| Weiser_2022b | Weiser et al. (2022) | 2,908 | 15 | ULS | RIEGL miniVUX-1UAV | temperate |
| Puliti_b | - | 621 | 3 | ULS | VUX1-UAV | boreal |
| FORinstance_NIBIO | Puliti et al. (2023) | 479 | 4 | ULS | RIEGL miniVUX-1UAV | boreal |
| FORinstance_SCION | Puliti et al. (2023) | 135 | 1 | ULS | RIEGL miniVUX-1UAV | temperate |
| FORinstance_CULS | Puliti et al. (2023) | 47 | 1 | ULS | RIEGL VUX-1UAV | temperate |

* corresponding to the dataset name in data published in Zenodo (DOI: 10.5281/zenodo.13255198)

As a result of such data source heterogeneity, the individual tree point clouds included in the FOR-species20K data exhibit a qualitative variation in terms of resolution, completeness (i.e. occlusion), and measurement accuracy (see Figure 2). Compared to multi-scan TLS data, the MLS and ULS point clouds were characterized by typical occlusions towards the top in MLS data or the lower parts of the canopy in ULS data (Schneider et al. 2019), whilst the single-scan TLS data shows both horizontal and vertical occlusion away from the scanner position.

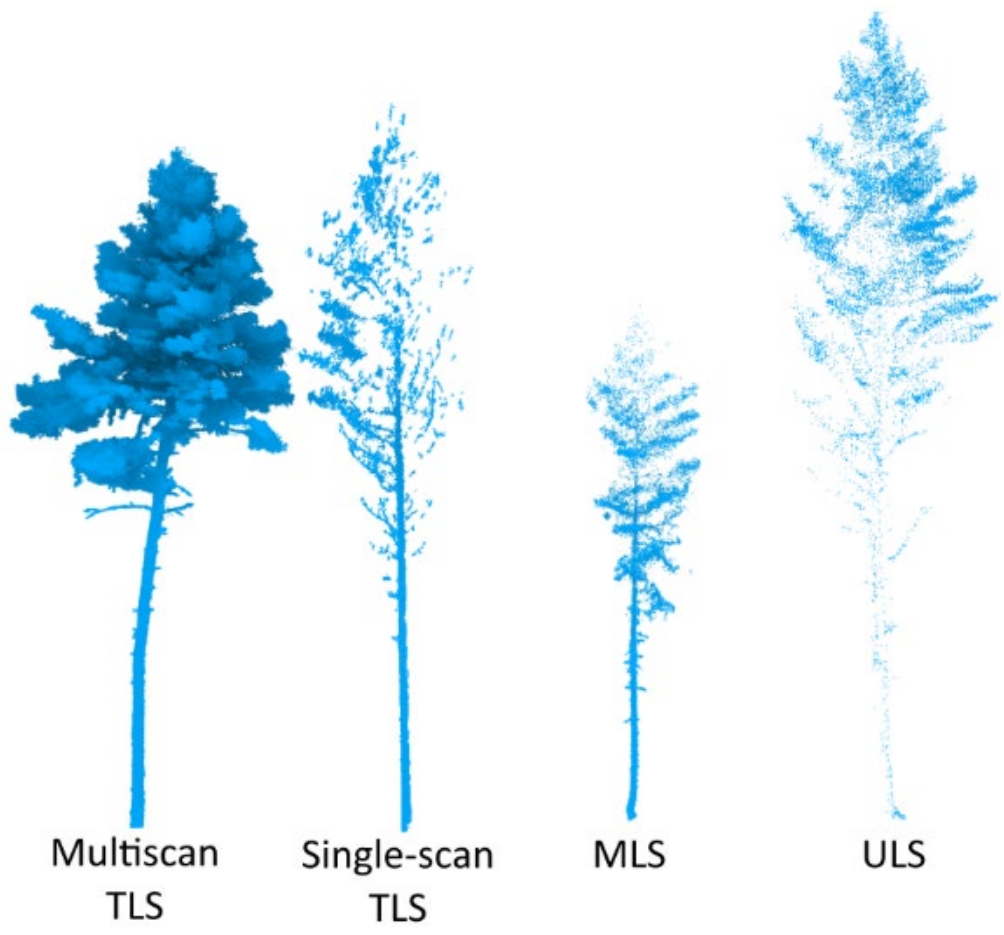

**Figure 2.** Example variation of the point cloud quality across platforms and different scanning protocols (TLS single- or multi-scan, MLS and ULS).

## 2.3 - Forest types and tree species

The FOR-species20K data covers the three main forest ecoregions in Europe with most of the trees from temperate forests (61%), followed by boreal forests (25%) and Mediterranean forests (7%). Further, a small percentage of trees was from temperate and boreal plantation forests outside Europe (4%) and tropical savannas (3%).

Compiling the FOR-species20K data, we retained only species with at least 50 individuals represented, resulting in 33 species of 19 genera (Figure 3). Whereas mostly focused on European forest tree species, FOR-species20K includes additional species from a more global ecological and climatic spectrum, including tree species from sclerophyll forests in Australia and broadleaf forests in North America. The most common species include *Pinus sylvestris,* with 3,296 samples, *Fagus sylvatica,* with 2,482 samples, and *Picea abies,* with 1,983 samples. Conversely, some species are much less represented, such as *Quercus robur* (195 trees), *Abies alba* (119 trees), *Larix decidua* (94 trees), with the rarest, *Prunus avium* having just 50 trees. While heavily imbalanced, this variation in species representation reflects realistic abundance distributions in European forest ecosystems with European dominant tree species being well represented and rarer species less so. FOR-species20K represents the most comprehensive dataset regarding the number of tree species openly available to date, which is essential for developing and evaluating robust classification models.

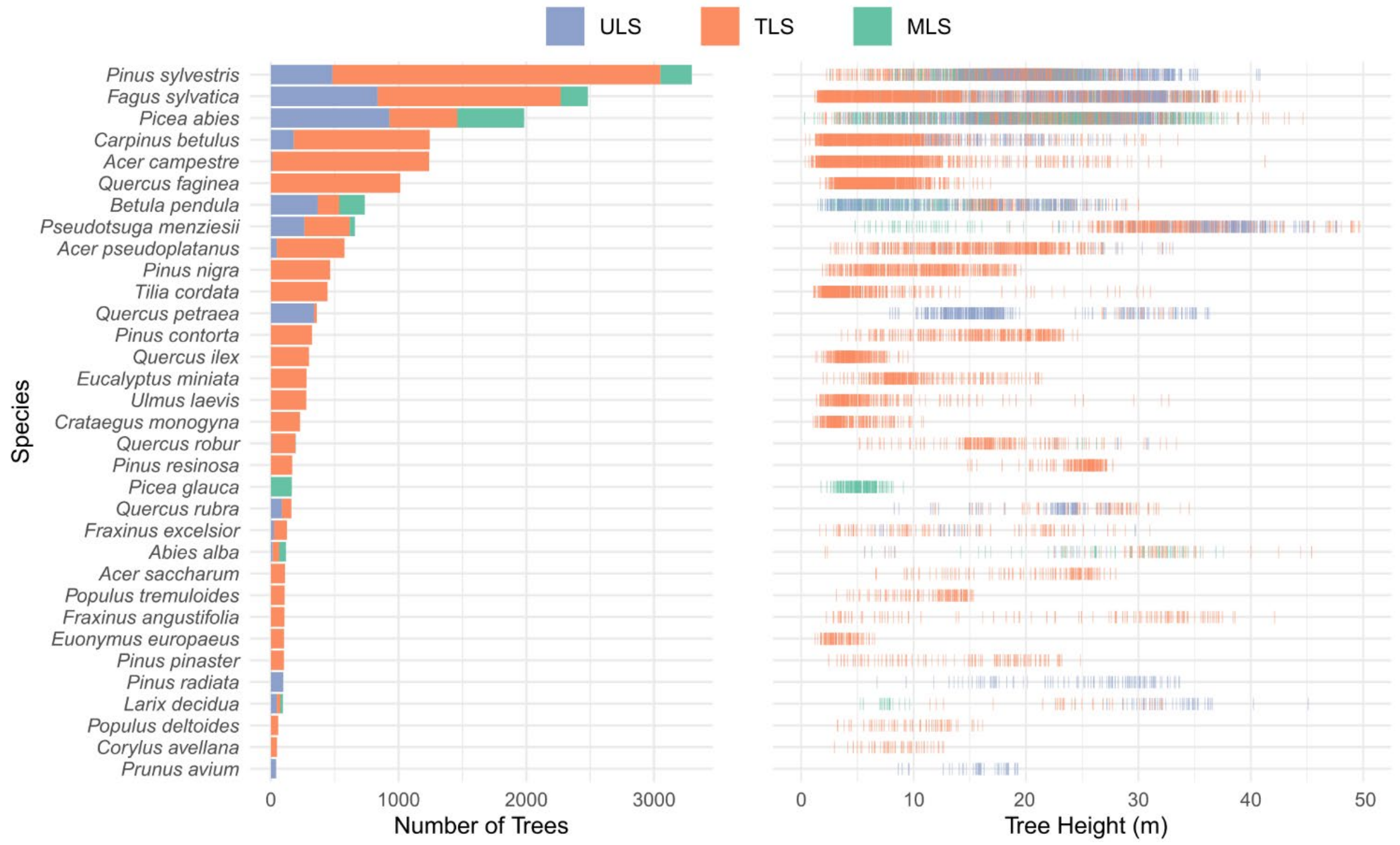

**Figure 3**. Tree species frequency and height distributions across data types in the FOR-species20K training data, split by sensor type.

## 2.4 - Dataset variability in tree size and crown architecture

The FOR-species20K dataset captures significant variation in tree height (Figure 3) and crown architecture (examples in Figure 4) encompassing a broad range of tree developmental stages under a variety of growth conditions.

Coniferous species cover a slightly wider height range (min = 0.3 m; mean = 20.4 m; max = 56.3 m, standard deviation=8.2 m) than broadleaf species (min = 0.3 m; mean = 11.4 m; max = 42.1 m; standard deviation=8.7 m). This variation in tree height also varies significantly according to species. For dominant European tree species, such as *P. abies*, *F. sylvatica*, and *P. sylvestris,* the dataset covered the full spectrum of forest developmental stages from young saplings to mature trees (see Figure 3). For rarer species such as *P. avium* the availability of data from only a few trees, often from the same stand, resulted in a rather uniform distribution of tree height (see Figure 3). Another source of variability in the FOR-species20K dataset is the significant intra-specific variation in crown architecture (see some visual examples in Figure 4), which is tightly linked to tree growth and competition under different growing conditions. Overall, the FOR-species20K dataset is highly diverse, encompassing a broad range of tree morphological variations. This diversity is a crucial feature for robust model training and evaluation in tree classification tasks.

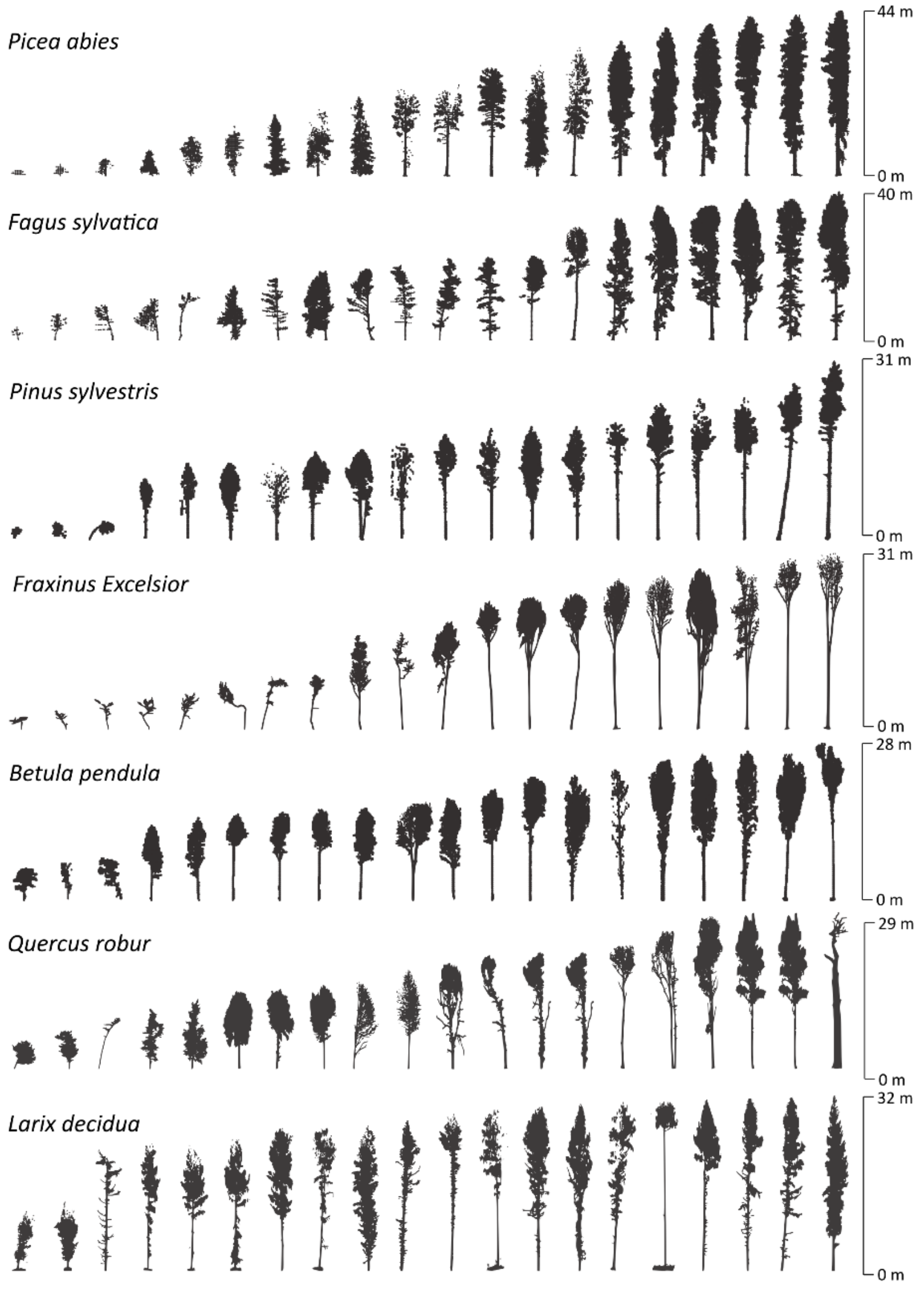

**Figure 4.** Examples of the variety of tree size and crown shape present in the FOR-species20K data for a selection of common European tree species.

## 2.5 - Data split

To benchmark different classifiers, the full dataset was split into development (90% of the trees; 17,707 trees) and test set (10% of the trees or 2,254 trees). The development set is composed of the individual tree point clouds and corresponding species labels and is meant to be used for both model training and validation. In the test set, the tree species labels were withheld from the participants to ensure a fair competition/unbiased evaluation

To address the challenge of benchmarking the classifiers' ability to handle largely imbalanced data in terms of tree species and size and data type, the test dataset was selected based on a stratified random sampling aimed at creating an artificially balanced dataset according to the following logic:

- **Balancing tree species**: to avoid overrepresentation of the dominant species, we set a maximum threshold of $n$=100 trees from each species

- **Balancing tree height**: to ensure a more balanced representation in terms of tree size and tree developmental stages, we defined a strata *A* consisting of i=20 tree height bins of 2.8 width.
- **Balancing data platforms**: to ensure a more homogeneous representation of the different platforms (ULS, MLS, and TLS) in the test data, we defined a strata *B*, where each statum is defined by the combination of a height bin from strata *A* with groups defined by the different platforms (i.e. $A_i$_*TLS*, $A_i$_*MLS*, $A_i$_*ULS*), resulting in a varying size *j* of strata depending of the availability of trees from different platforms across different height bins and species.
**Sampling:** for each species and strata *B*, we randomly sampled a number $n_B$ of trees, where $n_B$ is defined as the ratio between *n* and the *j* number of strata in *B*, thus ensuring the selection of a maximum of *n* trees for each species. If a species had fewer trees in a specific stratum than the target $n_B$, all available trees in that stratum were included in the sample.

# 3 - Methods

## 3.1 - Data science competition

Using the above-described dataset, a data science competition was launched in November of 2022 and lasted until June 2023 with the intention of benchmarking some of the most common classifiers available in the literature, including some of the latest model architectures. The competition was advertised in the same wide variety of channels as our call for data. Competitors tested seven different DL methods, including methods operating either directly on 3D point clouds or on 2D images obtained from projecting point clouds. Table 3 and the following subsections provide a high-level understanding of the benchmarked methods, their similarities, and their differences. Further implementation details can be found in Appendix 1.

**Table 3**. Summary of the main characteristics of the benchmarked DL architectures, including the input data, how the development data was split into training (train) and validation (val) data, the type of augmentation techniques adopted within the model, and specific characteristics related to the inference, or prediction step.

| Name | Data Input | train/val data split | Augmentation | Inference |
|---|---|---|---|---|
| **DGCNN + PointAugment** | 3D point clouds (4,096 pts tree$^{-1}$) | Tree size and tree species stratified 90% train; 10% val | PointAugment + downsampling, noise, rotation | Best model applied to test data |
| **Ensemble PointNet++** | 3D point clouds (8,192 pts tree$^{-1}$) | Same split as above | 6-fold rotation (z-Axis), random sampling | Ensemble classifiers, max avg prob. |
| **MinkNet** | Voxelized point clouds (>200 and <16,384 pts tree$^{-1}$) | Tree size and platform type stratified 90% train; 10% val | Random rotation along Z-axis | Majority voting after 50 rotations |
| **PointMixer** | 3D point clouds (4,098 pts tree$^{-1}$) | Tree size and platform type stratified 90% train; 10% val | None | Soft voting with 100 iterations |
| **SimpleView** | 6 x 2D projected images (512x512 pixels) | Random 90% train; 10% val | None | Accuracy on validation set |
| **DetailView** | 7 x 2D projected depth-images (256x256 pixels) | Weighted random sampler for the training data (98%), furthest distance sampling for the validation data (2%) | Point cloud random subsampling and rotation; Image flip | Averaged probabilities from 50 runs |
| **YOLOv5** | 4 x 2D projected images (600x800 pixels) | Tree species stratified 90% train; 10% val | YOLOv5 augmentations | Weighted mean of class probabilities |

### 3.1.1 Point-cloud based methods

*PointAugment and DGCNN (Brent Murray)*

This method combined PointAugment (Li et al. 2020), a generative adversarial network (GAN), and the Dynamic Graph Convolutional Neural Network (DGCNN; Wang et al. 2019). PointAugment enhances the point clouds to address species imbalance by generating more complex yet similar shapes. DGCNN, known for its ability to create and modify graph connections, aggregates these relationships for classification. Point clouds were down sampled to 4,096 points per tree, and manual augmentation was applied to balance the species. A species-wise stratified random subsample of 10% of the training trees was withheld from training and used as validation data for hyperparameter tuning and identification of the best model over the entire training period. The code for this method can be found at .

*Ensemble PointNet++ (Lukas Winiwarter)*

PointNet++ (Qi et al. 2017) was utilized to extract features directly from 3D point clouds through three sets of subsampling and grouping operations. Point clouds were down-sampled to 8,192 points per tree, and augmentations involved rotational augmentation (6-fold around the z-Axis) and random point sampling, improving prediction quality by using an ensemble of ten classifiers. The approach tested different configurations, finding that repeated random sampling and rotation significantly enhanced accuracy. The training involved various epochs based on the ensemble configuration and was done using the same split as the one used in the PointAugment and DGCNN method above. The code for this method can be found at https://github.com/lwiniwar/Tr3D_species_lwiniwar.

*MinkNet (Nataliia Rehush)*

MinkNet (Choy et al. 2019; https://github.com/NVIDIA/MinkowskiEngine) employs 3D sparse convolutions on voxelized point clouds using the Minkowski Engine framework. The model was calibrated with a 90% train, 10% validation split stratified by tree size and platform type and considering only trees with > 200 points, using a voxel size of 0.1 m. Training involved 250 epochs with data augmentation through random rotation along the Z-axis, resulting in a robust classification performance. During inference on the test data the model was applied to a sample 50 times while each time the point cloud was randomly rotated along the Z-axis. The final species was assigned based on majority voting (the most common label was chosen). The code for this implementation can be found at https://github.com/nrehush/minknet-tree-species.

*PointMixer (Hristina Hristova)*

PointMixer (Choe et al. 2022) blends features within and between point sets, making it effective for tree species classification. The training set was stratified to ensure a representative sample across different tree sizes and platform types. Using Farthest Point Sampling to 4,098 points per tree, the input points were selected and classified through soft voting. The training was set to 300 epochs and the best model (epoch 269) was selected using the validation overall accuracy. Inference was done using soft voting over 100 iterations with ten votes per iteration. The predicted probabilities for class labels were summed up and the final class label was chosen based on the highest sum value. The code for this implementation can be found at https://github.com/Hrisi/tree-species-classification.

### 3.1.2 Image based methods

All of the tested image-based methods relied on multi-view approaches, whereby the species classification is seen as an image classification task applied to different images of the same tree generated by projecting the point clouds onto 2D planes from different viewpoints.

### *SimpleView (Matt Allen)*

SimpleView (Goyal et al. 2021) uses a multi-view approach with six orthogonal camera projections of point clouds. Six projected images, coloured by depth, were used to train a ResNet-18 backbone (He et al. 2016). Previously applied to the tree species classification task by Allen et al. (2023), in this version, we modified the original repository (github.com/mataln/TLSpecies) to include larger images (from 256 × 256 pixels to 512 x 512 pixels), and down-sampling of the point clouds to 16,384 points. The model was trained on a simple random sample of 90% of the trees, and to account for data imbalance, the best model was selected as the one maximizing the validation (10% of the trees) balanced accuracy rather than the overall accuracy as in the original implementation.

### *DetailView (Julian Frey, Zoe Schindler)*

DetailView, builds upon SimpleView and incorporates dataset balancing by species, tree size, and platform through weighted random sampling. Further, DetailView uses a DenseNet-201 backbone (Iandola et al. 2014) and adds top and bottom views and a high-resolution projection of the trunk to leverage bark structure for classification. Only point clouds with at least 100 points were used for training. Image size was set to 256 x 256 pixels and augmentations included both point cloud (random subsampling and rotations) and image methods (flip) which were applied directly within the model. Further, tree size was included in the classification. The final predictions were obtained by averaging the 50 predicted probabilities per class and choosing the respective class with the maximum average probability. The code for this implementation can be found at https://github.com/JulFrey/DetailView.

### *YOLOv5 (Adrian Straker)*

The general concept of this approach is based on the application of a modified YOLOv5 architecture (Jocher et al. 2022) and the use of 4 side view orthographic projections (600 x 800 pixels) coloured by point count. We randomly split the training data set into 90 % for training and 10 % for validation. This split was conducted per species to ensure a representation of all species in both data sets. The final classification model was trained for 46 epochs using a stochastic gradient descent optimizer, a batch size of 128, and using default hyperparameters in the official YOLOv5 release. The model with the highest overall accuracy on the validation set was selected as the best model. The final tree species predictions were obtained by calculating a weighted average of the 20 predicted class probabilities per tree using weights of 0.5, 0.35, 0.05, 0.05 and 0.05 for the highest five predicted class probabilities per side view image, respectively and selecting the class with the highest weighted average probability. The code for this method can found at https://github.com/AWF-GAUG/Yolov5-tree-species-classification-in-point-cloud-images.

## 3.2 - Benchmarking and metrics

The presented models were benchmarked in relation to their ability to classify tree species in the unseen test data. The ground truth (GT) and predicted tree (PT) species were used to generate the confusion matrices based on which we obtained the counts for the true positives (TP), false positives (FP), and false negatives (FN) required to compute a selection of commonly used metrics for the evaluation of classification tasks. The following metrics were used to evaluate the species-wise and global model's predictive accuracy:

$$Overall\ accuracy\ (OA) = \frac{TP}{GT} \quad \text{(Eq. 1)}$$

$$Species\ accuracy\ (SA_s) = \frac{TP_s}{GT_s} \quad \text{(Eq. 2)}$$

$$Precision\ (P) = \frac{TP}{TP + FP} \quad \text{(Eq. 2)}$$

$$Recall\ (R) = \frac{TP}{TP + FN} \quad \text{(Eq. 3)}$$

$$F1 - score = \frac{2 \times P \times R}{P + R} \quad \text{(Eq. 4)}$$

Further, to address the ability of the model to get a fully agnostic understanding of the task we evaluated the overall accuracy for the different data types (TLS, MLS, and ULS), and performance across different tree sizes.

For future benchmarking against the FOR-species20K test data, we established a Codabench (Xu et al. 2022) benchmarking page which will be made openly available in the future.

# 4- Results and discussion

## 4.1- Benchmarking

The leaderboard for the results of the data science competition is shown in Table 4. The best performing model by overall accuracy (OA), recall and F1-score was DetailView, whilst the best performing model by precision was YOLOv5, with DetailView a close second. We therefore interpret DetailView as the best performing model overall. We found a notable performance disparity between multi-view image-based methods (average OA = 77.8%) and point cloud methods (average OA = 72.1%), with the worst-performing image-based method outperforming the best-performing point cloud method.

**Table 4**. Leaderboard for the data science competition described in this study, ordered by overall accuracy. The bold values represent the best results on the FOR-species20K test dataset for each metric.

| Tested method | Input data | Overall Accuracy (%) | Precision (%) | Recall (%) | F1 (%) |
|---|---|---|---|---|---|
| DetailView | Image | **79.5** | 82.3 | **76.7** | **78.0** |
| YOLOv5 | Image | 77.9 | **84.2** | 75.0 | 77.3 |
| SimpleView | Image | 76.2 | 76.9 | 75.5 | 75.6 |
| Ensemble PointNet++ | Point cloud | 75.6 | 78.2 | 73.5 | 74.9 |
| MinkNet | Point cloud | 73.7 | 79.9 | 70.6 | 72.3 |
| PointMixer | Point cloud | 71.1 | 74.4 | 65.5 | 71.1 |
| PointAugment + DGCNN | Point cloud | 68.3 | 72.5 | 65.7 | 70.3 |

The performance disparity between image and point cloud methods may be due to the current technology readiness level of the two methods, with image-based approaches benefitting from the matured feature extraction capabilities of CNNs and traditional image processing techniques such as extensive data augmentation and the simplification of data through 2D projections. These techniques simplify the input data structure by making it uniform (pixel grid) and allow for the projection of many more points than allowed by point cloud methods thus allowing for a more efficient data compression and digestion by the model. Another potential advantage of multi-view CNNs is that the final prediction is derived from an ensemble of predictions across different projections, which may enhance the accuracy and reliability of the results.

In contrast, point cloud-based methods encountered challenges stemming from the sparse and unstructured nature of the 3D data. Although architectures like PointNet++ and DGCNN are designed to tackle these problems, they are relatively recent developments and continue to face difficulties related to high computational demands and, therefore, the need to heavily subsample the original point clouds. Such information loss is likely one of the causes of the poorer performance of the point cloud methods. On the other hand, we expect point cloud methods to have a larger potential for improvements over time, since they are starting from a lower baseline. Thus, although we found that image-based methods represent the current state-of-the-art in terms of classification performance, it is likely that point cloud architectures will attract more development and may even outperform 2D methods in the future.

The confusion matrix of the best performing method, DetailView (Figure 4), shows that, on average, the overall accuracy for coniferous species (87.4%) was larger than for the broadleaved species (71.3%). Such a result is likely influenced by the larger number of broadleaf species (22 species) compared to only 11 coniferous species, as it is typical in classification tasks that performance drops when the number of classes increases. In addition, broadleaved trees typically develop more plastic and asymmetric crowns which substantially increase the intra-specific variation in tree structure and thus the ability of finding common representative features within a species. Conversely, the better performance for coniferous trees may be seen as surprising when taking into account that 7 of the 11 species are from the same genus (*Pinus*). While we found examples of intra-genus confusion (e.g. between *P. pinaster* and *P. nigra* or between *Q. faginea* and *Q. ilex*), these were not noticeably more frequent or severe compared to the confusion amongst species in different genera.

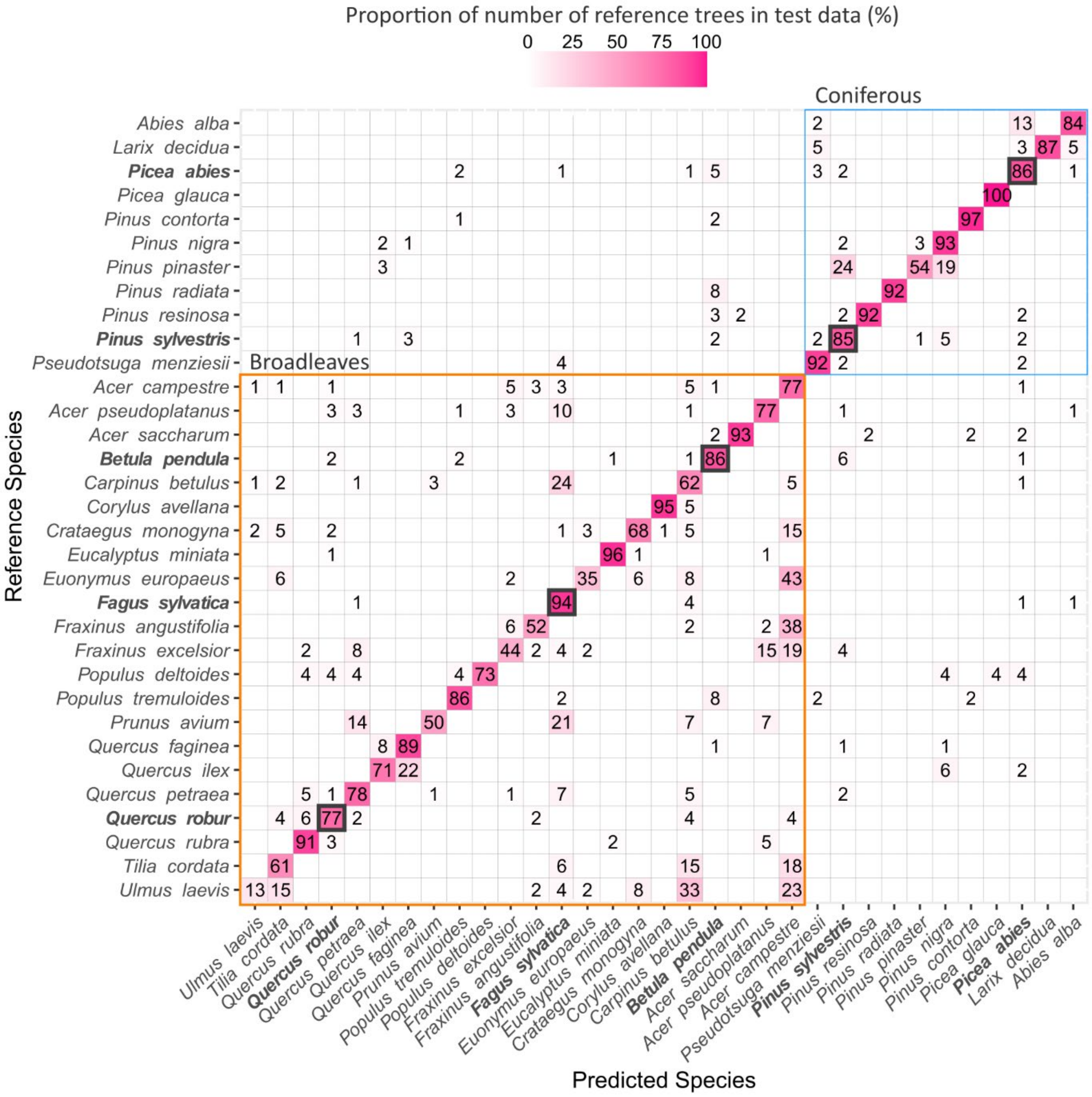

**Figure 4**. Normalized confusion matrix (%) computed using the DetailView method on the withheld test data. The confusion matrix is ordered alphabetically by coniferous and broadleaved species to allow within- and cross-genus comparison. The boxes with black outline highlight the performance on some of the most abundant species in European forests (in bold).

We examined the influence of data imbalances across tree species and datasets on classification accuracy (Figure 5) and found that neither the quantity of data points nor the number of datasets within each species significantly affected the model F1-score (with Pearson's correlation coefficients being -0.03 and 0.06, respectively). While this finding may have been partly driven by a degree of similarity between the development and test trees, it suggests that the DetailView model effectively compensates for data imbalances.

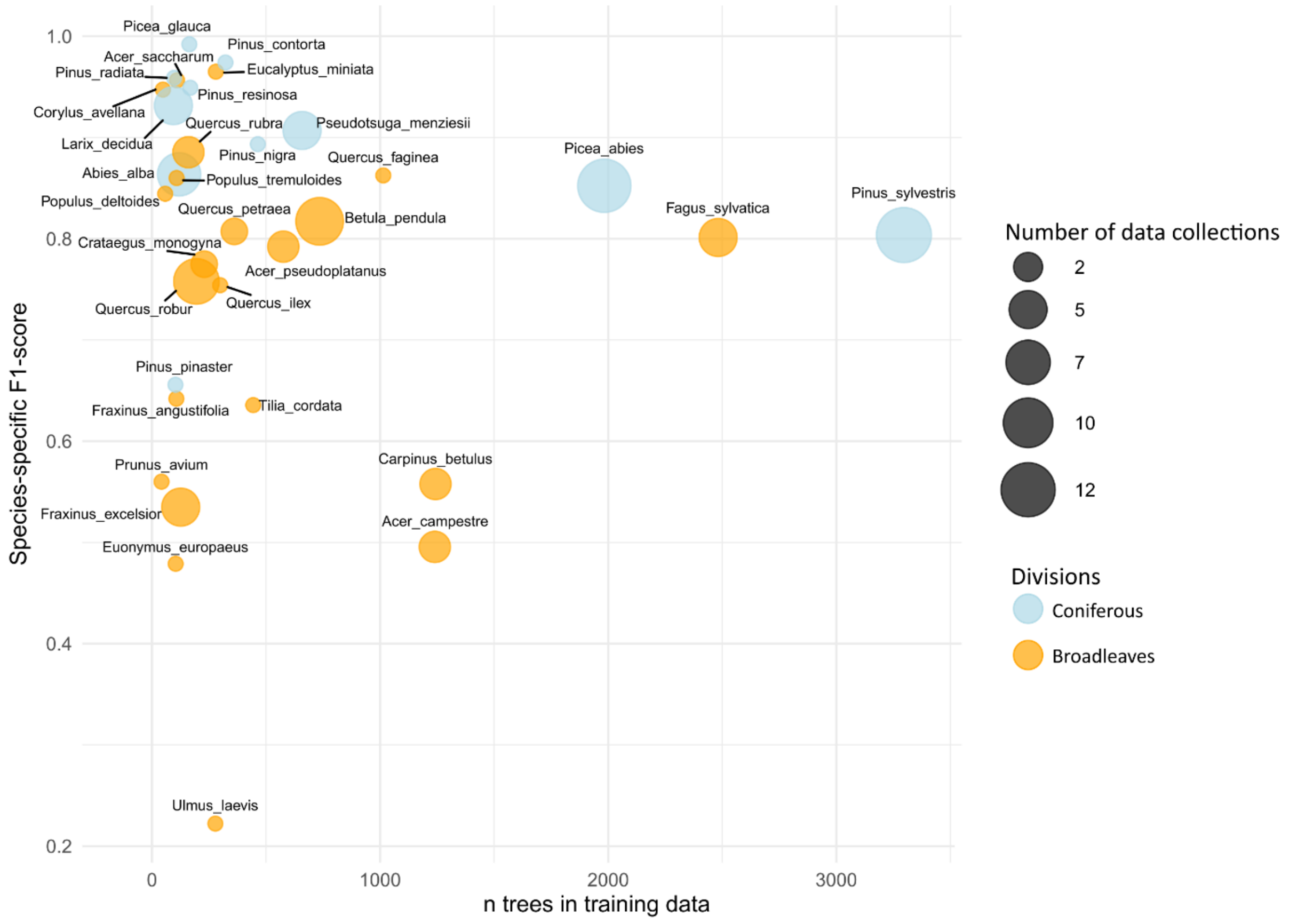

**Figure 5**. Bubble plot representing the relationship between the sample size in the training data and the F1-score obtained in the test data for each species. The size of the bubbles represents the number of datasets that contributed to each of the species. The plot was computed using the results from the best performing method (DetailView).

In addition to the above results, a visual analysis of a sample of the incorrectly classified trees (see examples in Figure 6) suggests that the model's performance may influenced by the lack of crown architectural clues (e.g. branches and branching patterns) in small trees. These trees often belong to lower layers in the canopy, where significant competition strongly affects the shape and complexity of tree crown architecture. Thus, the model's performance issues seem to stem more from these architectural complexities and incomplete crown data rather than the sheer volume of data available.

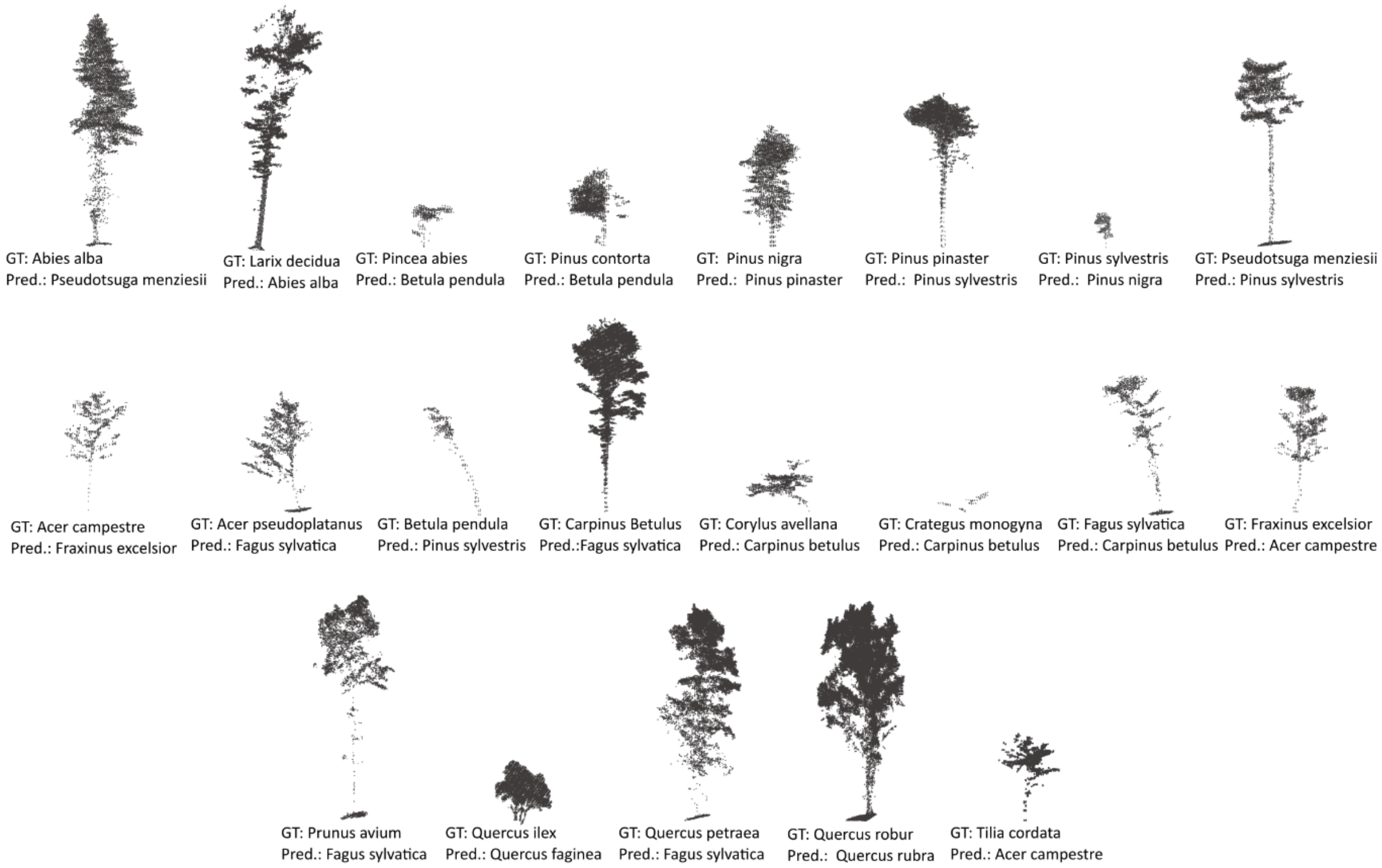

**Figure 6**. Examples of misclassified trees for some tree species of forestry relevance in Europe, comparing ground truth (GT) and predicted (Pred.) species using the DetailView method.

## 4.2- Accuracy by acquisition platform

The assessment of the platform-agnostic capabilities of the tested models (see Figure 7) revealed that for the top five methods, the overall accuracy >70% for all platforms. OA was highest for all methods for MLS data, and DetailView performed best for MLS and ULS data, but was outperformed by 0.1% for TLS data by YOLOv5.

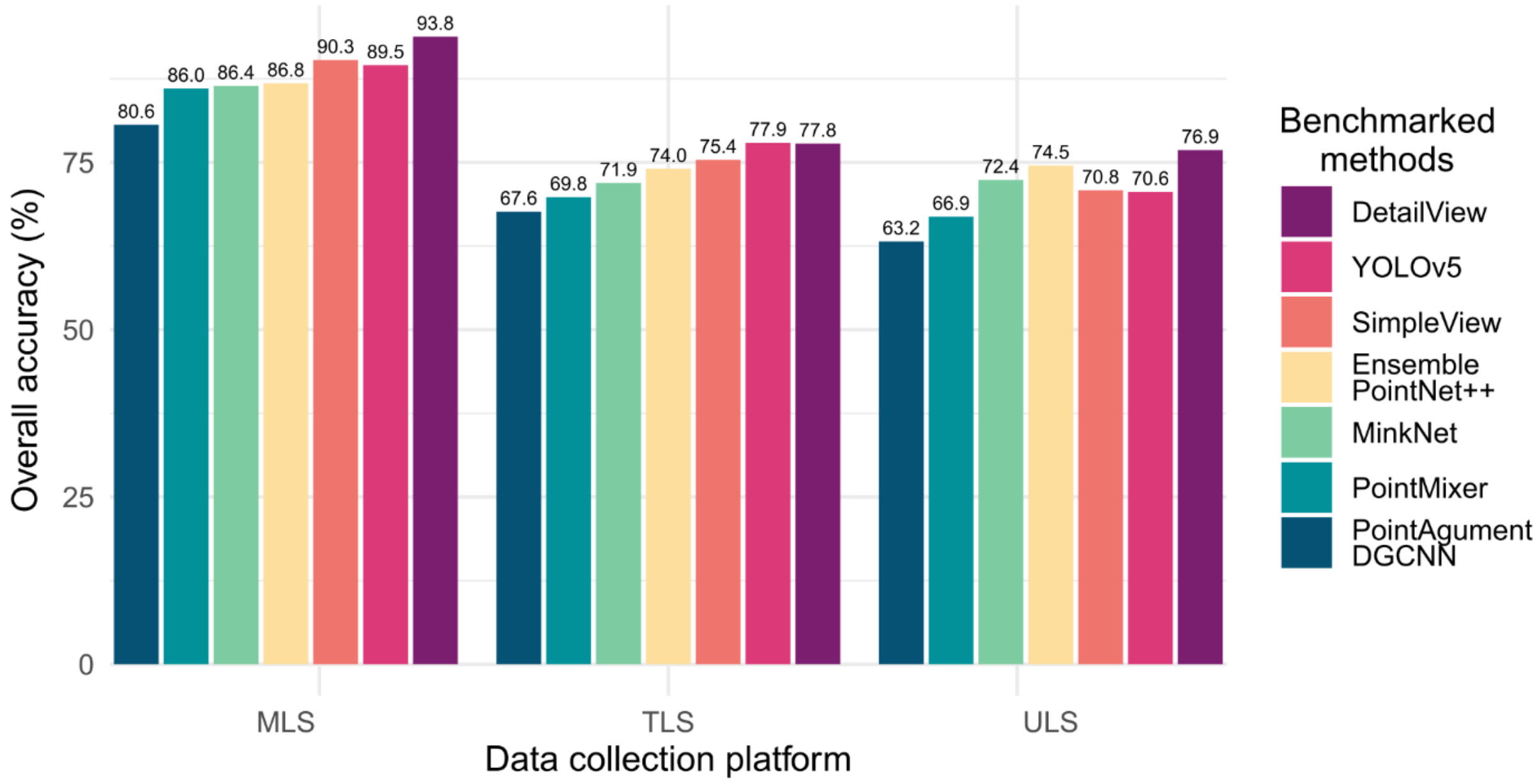

**Figure 7**. Overall accuracy by benchmarked method for the tree laser scanning data collection platform: mobile (MLS), terrestrial (TLS), and uncrewed aerial vehicle (ULS).

The above result shows that not only is tree species classification reliable from each platform, but that all models performed well regardless of the laser scanning data source. This capability is particularly relevant for streamlining the adoption of tree species classifiers in operational settings. By unifying under a single model, users can avoid the complexity of adopting different processing pipelines for the autonomous characterization of forest environments using proximal laser scanned data. Additionally, this ensures consistency in outputs across different data modalities, which is crucial as multiple technologies are increasingly integrated into forest information systems.

Although the experimental results (see Figure 7) generally show higher predictive accuracy for MLS data, these values are likely inflated because MLS data represent only a small portion of the FOR-species20K dataset, both by the number of sample trees and species diversity (just 8 species have MLS data, vs 30 for TLS; Figure 3).

Interestingly, for the sparser ULS data, the PointNet++ ensemble and the Minknet models outperformed YOLOv5 and SimpleView. This suggests that point cloud-based methods may better leverage sparser point representations and face challenges with very dense datasets, or alternatively image methods may struggle where there is less information in the data. Point cloud-based methods often reduce computational demands by subsampling the point cloud, which consequently reduces the richness of the information. The superior performance of DetailView on the rarer MLS and ULS data might be explained by the weighted random sampler used in DetailView (see description in section 3.1.2), which allows balancing the training and validation data specifically based on the platform type.

## 4.3- Accuracy by tree size

The analysis of the impact of tree size on DetailView's overall accuracy showed that for trees taller than around 8 m the accuracy remained relatively stable regardless of tree height (Figure 8). However, accuracy was substantially lower for small trees (< 5 m in height), falling as low as 25% for trees shorter than 2 m.

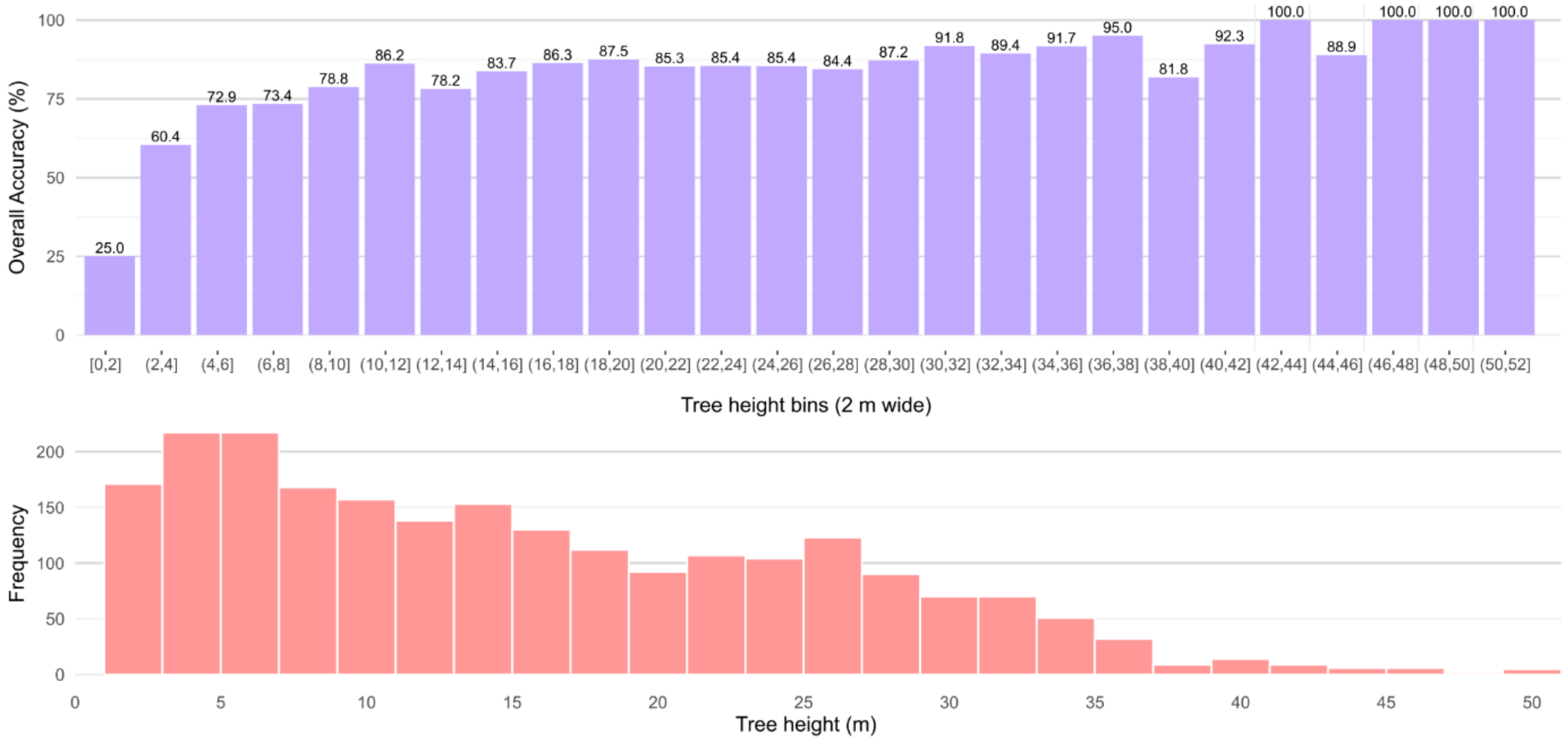

**Figure 8**. DetailView's overall accuracy (%) categorized by tree size with tree heights binned into 2-meter intervals.

## 4.4- Domain/application evaluation

Although the primary aim of the FOR-species20K dataset was to objectively benchmark different species classifiers, rather than providing a complete tree species database for training models for operational tree species prediction, we examined three DetailView's key potential application scenarios in mature forest (tree height > 5 m) in three European biogeographic regions, including boreal, hemiboreal, and temperate regions (Table 5). Unfortunately, due to the lack of enough sample trees for Mediterranean species such scenario was here omitted.

**Table 5**. DetailView's overall accuracy for three European biogeographic regions with increasing species diversity and across different input data types.

| Biogreographic region | Available tree species | Input data type | n test trees | Overall accuracy (%) |
|---|---|---|---|---|
| Boreal | *Picea abies*, *Pinus sylvestris*, *Betula pendula* | ULS | 87 | 90.8 |
| | | TLS | 100 | 87.0 |
| | | MLS | 90 | 88.9 |
| Hemiboreal | *Picea abies*, *Pinus sylvestris*, *Betula pendula*, *Quercus robur*, *Fagus sylvatica*, *Fraxinus excelsior*, *Acer pseudoplatanus*, *Corylus avellana* | ULS | 167 | 79.0 |
| | | TLS | 280 | 84.3 |
| | | MLS | 138 | 92.8 |
| Temperate | *Picea abies*, *Pinus sylvestris*, *Betula pendula*, *Quercus robur*, *Fagus sylvatica*, *Fraxinus excelsior*, *Acer pseudoplatanus*, *Corylus avellana*, *Pseudotsuga menziesii, Abies alba, Larix decidua, Carpinus betulus, Quercus petraea, Prunus avium, Populus tremuloides, Crataegus monogyna*, *Tilia cordata* | ULS | 367 | 75.7 |
| | | TLS | 517 | 82.8 |
| | | MLS | 179 | 92.7 |

While the above analysis is valid only for the  We found that DetailView performs reliably across different biomes, consistently achieving high overall accuracy (> 75%) with all data platforms in all ecoregions. DetailView's accuracy was highest (87-90%) in species-poor biomes, such as boreal forests, and decreased as tree species richness increased, as seen in the transition from predominantly coniferous boreal forests to mixed broadleaved temperate forests.

## 4.5- Limitations

While these results are promising for the operational use of the DetailView model, users must be aware that these results are likely inflated due to the use of a test dataset split from the same pool of data as the training dataset and thus are assumed to have similar data properties and tree morphology. We suggest users keep in mind the following issues when deploying any model trained on FOR-species20K to new datasets:

- Open-set recognition: The species list is incomplete even for Europe, leading to incorrect predictions for species unseen to the model and likely impacting the real-world performance in diverse forests encompassing species beyond those available in the FOR-species20K data. Therefore, the models published here should only be applied to data with known species diversity within the FOR-species20K diversity.

- Differences in quality of the tree segmentation: FOR-species20K is composed of very high-quality tree segmentation, and thus it is unclear to what extent the model can be applied to poorly segmented trees. This is particularly relevant for complex forest structures where individual trees are difficult to accurately segment due to multiple canopy layers and overlapping crowns.
- Species representation on different platforms or sensors: Some species have been recorded by a single platform only, and our dataset uses a relatively small range of sensors (Table 2), and model performance on other platforms or sensors is untested.

## 5 - Conclusions

This study, addressing the need for benchmarked individual tree species classifiers, is part of a broader effort to accurately characterize forest ecosystems using automated methods relying on proximal laser scanning data. In this context, DL tree species classification models are a key component in larger DL pipelines that also includes individual tree segmentation. While FOR-species20k represents an important starting point in bringing together the scientific community to build critical data infrastructure for benchmarking and developing species classification from proximal sensed laser scanning data, enabling the development of the next generation of species classifiers requires a community effort, including further expanding the database. Here, the focus should be on extending the dataset with more tree species, increasing the sample size for poorly represented species and their seedlings, as well as increasing the number of trees from underrepresented scanning approaches (i.e. MLS and ULS), and tree size classes.

## Acknowledgements

This work was supported by the COST Action 3DForEcoTech (CA20118). This work is part of the Center for Research-based Innovation SmartForest: Bringing Industry 4.0 to the Norwegian forest sector (NFR SFI project no. 309671, smartforest.no). ERL and HJFO were funded by a UKRI Future Leaders Fellowship awarded to E.R.L. (MR/T019832/1). MJA was supported by the UKRI Centre for Doctoral Training in Application of Artificial Intelligence to the study of Environmental Risks (EP/S022961/1). We acknowledge the technical support and compute time at the Vienna Scientific Cluster VSC-5 for parts of the Ensemble-PointNet++ results. LW was funded in part by the Austrian Science Fund (FWF) [J4672]. The contribution of the DetailView model was funded by the Deutsche Forschungsgemeinschaft (DFG, German Research Foundation) – Project FR 4404/1-1. NS was supported by the Academy of Finland through UNITE Flagship (357906) and Scan4rest Research Infrastructure (346382). KC was funded by the European Union (ERC-2021-STG Grant agreement No. 101039795). ET was funded by INEST - PNRR (Italian National Plan for Recovery and Resilience), Project id, ECS00000043. REMBIOFOR dataset was funded by National Centre for Research and Development in Poland underthe BIOSTRATEG programme (grant agreement number BIOSTRATEG1/267755/4/NCBR/2015), project REMBIOFOR 'Remote sensing-based assessment of woody biomass and carbon storage in forests'.

Views and opinions expressed are however those of the author(s) only and do not necessarily reflect those of the respective funding agencies which can therefore not be held responsible for them.

## Data Availability statement

The FOR-species20K data are openly available at https://zenodo.org/records/13255198 and the code for the different methods benchmarked in this study can be found at the following GitHub repositories:

| Benchmarked method | GitHub repository |
|---|---|
| PointAugment and DGCNN | https://github.com/Brent-Murray/TR3D_PointAugDGCNN |
| Ensemble PointNet++ | https://github.com/lwiniwar/Tr3D_species_lwiniwar |
| MinkNet | https://github.com/nrehush/minknet-tree-species. |
| PointMixer | https://github.com/Hrisi/tree-species-classification |
| SimpleView | https://github.com/mataln/TLSpecies |
| DetailView | https://github.com/JulFrey/DetailView |
| YOLOv5 | https://github.com/AWF-GAUG/Yolov5-tree-species-classification-in-point-cloud-images |